\pdfoutput=1
\documentclass[11pt]{article}
\usepackage{graphicx,amsmath,amssymb,amsfonts,amsthm,xspace}
\usepackage{cmll}

\usepackage{acl}

\begin{document}
\title{Pronoun Logic} 
\author{Rose Bohrer \and Ashe Neth\\
        Computer Science Department \\
        Worcester Polytechnic Institute \\
        Worcester, MA, USA \\
        \texttt{\{rbohrer,aneth\}@wpi.edu}}

{\makeatletter\acl@finalcopytrue
  \maketitle
}              

\begin{abstract}
Particularly in transgender and nonbinary (TGNB) communities, it is an increasingly common practice to publicly share one's personal pronouns so that we may be gendered correctly in others' speech.
Many of us have nuanced desires for how we are gendered, leading us to use more complex descriptions of our wishes; for example, the descriptor  `she/they'.

We observe that these descriptions of our wishes have the structure of a little language all their own.
We thus propose formal logic as a tool for expressing one's personal pronouns and potentially other aspects of gender.
We explore three potential logical foundations (linear logic, temporal logic, and free logic with definite descriptions) and  their trade-offs.

Our foremost motivation for this proposal is play, affirming that one can be both a logician and TGNB at the same time.
We present formalization as something that can continue to evolve over time with society's understanding of gender.
This implies that outreach is a major potential application: we can show TGNB youth that they belong in logic and have a unique contribution to make.
Tools for evaluating whether one's pronouns are respected are an application as well.
\end{abstract}

\newcommand{\pron}[1]{\textit{#1}\xspace}

\section{Introduction}

A \emph{personal pronoun} is a pronoun whose referent\footnote{Throughout the paper, we use \emph{referent} to mean the person who is spoken about and \emph{speaker} for the one who speaks.} is a person.
When a transgender or non-binary person is the referent, many speakers fail to infer the referent from context, and even more often, fail to infer the pronoun by which to refer, i.e., they \emph{misgender}\footnote{\emph{Misgendering} includes using incorrect pronouns or gendered language on an individual. It also includes \emph{deadnaming}, the use of a former name changed for purposes of gender affirmation.} the referent.

In many contemporary communities, particularly online communities, it has become common practice to pair ones name with a \emph{pronoun descriptor} (PD) indicating which personal pronouns should be used to refer to oneself, in order to (1) reduce the occurrence of misgendering and (2) among cis people, express solidarity to transgender and non-binary (TGNB)\footnote{This term is not meant to imply that transgender and non-binary are mutually exclusive categories. There are both nonbinary transgender people and nonbinary people who are not transgender.} people.
This matters because using one's correct gender has been linked to feelings of acceptance and even progress in civil rights~\cite{baron2020s}, yet people outside of TGNB communities may have limited experience with communicating one's linguistic expectations~\cite{pewReport}.

Pronoun descriptors not only describe language, but follow a language of their own. Base descriptors such as \pron{she/her}, \pron{he/him}, and \pron{they/them} indicate an individual with exactly one pronoun class, containing an object and subject case.

More complex pronoun descriptors exist, however, especially among nonbinary people. 
For example, the descriptors \pron{they/she} and \pron{she/they} both indicate that both \pron{they/them} and \pron{she/her} pronouns are correct references to an individual.

This ability to \emph{compose} pronoun descriptors suggests the latent structure of a formal logic, yet such structure has yet to be studied, to the best of our knowledge.
This is a gap in the literature: not only is formal logic a key tool in symbolic AI broadly, but it is intimately interrelated with the formalizations of natural language that make computational linguistics possible in particular.

Because this topic is so unstudied, the present paper sets a modest goal, to explore how different logical foundations might be used to formalize pronoun descriptors, and in so doing, lay out a design space for such a logic.
Our goal is to reveal the act of formalization as an act of gender expression, through which TGNB people in logic might announce which aspects of gender are most important to us personally.

\subsection{Context: Structuralism}
Resistance to structuralist universals is often seen as central to queer theory.
This is in tension with the central use of public labels such as pronoun descriptors in trans and nonbinary communities: pronouns represent a contract which members of a community are expected to uphold.

This paper is meant as a provocation which uses pronouns descriptors as a means to pull at the tension between structural and post-structural views of gender.
We outline three potential logical formalisms for pronoun descriptors.
In doing so, we explore the idea that, although no one formalism will capture every nuance of gender identity, formalism is just as socially constructed as gender, and just as gender, the development of formalism can be viewed as a creative act.
Through this exploration, we provoke readers to view formal models of gender, specifically of gender pronouns, as something which can co-evolve with society's understanding of gender, not something to be avoided.

The following sections explore three potential logical foundations.
The first two foundations are linear logic~\cite{DBLP:journals/tcs/Girard87} and temporal logic~\cite{pnueli1977temporal}, both applied directly to formalization of PDs.
The third proposed foundation is free logic with description operators~\cite{DBLP:conf/cade/BohrerFP19,sep-logic-free}, which allows us to begin moving beyond pronoun descriptors to broader descriptions of gender.
Each proposal makes it easier to formalize different aspects of gendered communication, reflecting different priorities which might be held by different TGNB people.
Linear logic makes it easy to clarify \emph{who} chooses which pronoun to use.
Temporal logic makes it easy to set requirements on a pattern of speech that emerges \emph{over time}.
Description operations can formalize the desire to go beyond basic pronoun labels, to labeling ourselves based on what predicates we satisfy, i.e., \emph{what kind of person we are}.
In closing with description operations, we invite broader exploration of formalizing experiences of gender beyond pronoun labels.

\subsection{Positionality Statement}
The authors are native English speakers working in the United States.
We work in logic-based approaches to symbolic AI.
Pronouns and other references to gender vary wildly across languages and communities; we recognize that the approach taken here is Anglo-centric.
The authors ourselves are also speakers of Mandarin Chinese, Japanese, and Korean, and are excited for future work which embraces the radically different roles of gender pronouns across languages.
Examples include the gender-neutrality of third-person pronouns in spoken Mandarin and the minimal use of gendered third-person pronouns in Japanese, where gender expression occurs far more often through first-person pronouns.

The first author is a binary transgender woman.


%



\section{Linear Logic} 
Linear logic~\cite{DBLP:journals/tcs/Girard87} is a version of logic where every assumption must be used exactly once.
This decision has deep implications on linear logic's expressive power, resulting in a logic that is well-suited for modeling consumable, ownable resources and which features multiple notions of choice.
In this section, we focus on the choice aspect of linear logic, leaving exploration of pronouns-as-resources to future work.

The option of linear logic as a foundation for Pronoun Logic is motivated directly by a real phenomenon observed in the authors' lives.
Many people, especially nonbinary people, may use \emph{dual pronouns}, they may have two (or more) distinct sets of pronouns which apply to them.
For example, one might write the descriptor \emph{she/they} to indicate that both \emph{she/her} and \emph{they/them} pronouns are applicable.
This represents a \emph{choice} between different pronoun sets.
Herein lies an ambiguity: who gets to decide?
Most often, the speaker decides, but this is not always the case.
Some (e.g., perhaps genderfluid) people may have a varying innate sense of gender from day to day, and thus the best pronoun for that person may vary from day to day as well, with only them being able to decide.
Thus, we wish to model two notions of choice, one where the speaker chooses and one where the referent chooses.
As we will see, linear logic is well-suited.

We present a proposed (fragment of) linear pronoun logic, where the primary syntactic class (formulas) stands for pronoun descriptors (as descriptions of what the speaker understands to be the referent's true pronoun descriptor).
We define pronoun descriptors $A,B,C$ by the grammar:

\[ A,B,C {:}{:}{=}  s/o ~|~ A\ \&\ B ~|~ A \oplus B ~|~ A \otimes B~|~A \multimap B \]

We describe each (social) construct in turn, comparing its proposed meaning in linear pronoun logic with its standard meaning in linear logic.
The most common atomic PDs are the subject-object (s/o) PDs.
The best-known s/o PDs are \pron{she/her}, \pron{he/him}, and \pron{they/them}.
As of this writing, others in use include \pron{ze/zir}, \pron{hir/hir}, \pron{hy/hym}, \pron{ze/zem}, \pron{it/it}, and \pron{vae/vem}.
However, there is no reason to assume the set of s/o PDs is less than countably infinite; thus we treat it as an arbitrary countable set.
Subject-object pronouns correspond to propositional letters in linear logic, i.e., they function logically as arbitrary names.

The most important composite PD is the \emph{external choice} $A\ \&\ B,$ which indicates that the speaker has the choice between $A$ and $B$.
This corresponds directly to external choice in linear logic, where $A$ and $B$ must both be provable from the same assumptions, so that a speaker can choose which one to use.
The \emph{internal choice}, written $A \oplus B,$ means that the referent chooses whether to be referred to by pronoun descriptor $A$ vs.\ $B$.
This corresponds directly to internal choice in linear logic, where only one of $A$ or $B$ need be producible from the given assumptions, and the speaker is forced to work with whichever one was proven.
Having two distinct choice operators allows us to draw crucial distinctions in pronoun usage patterns.
The pronoun descriptor $\pron{he/him} \oplus \pron{she/her}$ is radically different from $\pron{he/him}\ \&\ \pron{she/her}$. 
The former means that the referent decides in which contexts they will be called he vs.\ she, the other means the speaker decides.
To see how important this distinction is: consider as an example a trans man who is out to coworkers but not family.
He would likely choose \pron{he/him} around coworkers and \pron{she/her} around family, but it is disrespectful misgendering if coworkers use \pron{she/her} and runs risk of outing if \pron{he/him} is used with family.
Thus this person's scenario deserves being distinguished from $\pron{he/him}\ \&\ \pron{she/her}$ where the speaker may choose freely.

The product $A \otimes B$ indicates that both $A$ and $B$ must be used.
This can be important for a person who identifies confidently with multiple pronouns.
Consider a nonbinary masculine person using  $\pron{he}/\pron{they}$ pronouns.
If he wishes to indicate that both their masculine and nonbinary traits ought to be affirmed, they can do so with $\pron{he/him} \otimes \pron{they/them}$.
An astute reader might question whether $A \otimes B$ and $B \otimes A$ ought be considered the same or not, and this question is just as subtle for pronoun logic as for pronoun descriptors in the wild or for linear logic.
In linear logic, $\otimes$ is commutative, as $A \otimes B$ and $B \otimes A$ are interderivable, yet $\otimes$ is often imbued with a left-to-right operational semantics, wherein their \emph{operational behavior} is distinct!
A similar nuance applies to pronoun descriptors in the wild, where any utterance which respects $A \otimes B$ simultaneously respects $B \otimes A$ pronouns.
Yet, in practice, many people imbue their descriptors with additional semantics.
If someone updates an email signature from $\pron{they}/\pron{she}$ to $\pron{she}/\pron{they},$ it may be intended to announce a change in preference, that both pronouns are still considered respectful but that $\pron{she/her}$ has taken preference. Just as in operational semantics of linear-logic, a left-to-right reading tells us an added semantics, i.e., which one ``comes first.'' This is to say that while order may not strictly matter, in the case of a human interpreting and writing a PD, order may have some impact on the intended and/or inferred meaning which cannot be described formally, marking a limitation of this logic.

How should we interpret $A \multimap B$ for pronouns? 
This is perhaps the least clear of all the connectives.
To lend insight, we review the interpretations (semantics) that previous scholars have used for linear logic.
Two well-established interpretations are message-passing interpretations~\cite{DBLP:conf/csl/DeYoungCPT12} and game-theoretic interpretations~\cite{DBLP:journals/apal/Japaridze97}.
In a message-passing interpretation, $A \multimap B$ means ``I receive $A$ then send $B$'' and in a game-theoretic interpretation it means ``my opponent does $A$, then I do $B$.''

Recall that in our interpretation, PDs $A$ are interpreted from the referent's view, e.g., internal choice $\oplus$ means referent's choice.
Thus we interpret $A \multimap B$ to mean ``the speaker says $A$, then the referent replies $B$.''
Though the authors are not aware of anyone who uses such a PD, this notation might be used to model common transphobic accusations against TGNB people.
Many TGNB people, especially many NB people, have witnessed (false) accusations that they are ``doing it for the attention,'' and as an extension of this, some might face actions that they ``just want to correct people.''
The linear implication operator $A \multimap B$ allows us to model such a hypothetical want: $\pron{he/him} \multimap \pron{they/them}$ means ``I specifically want you to call me `he' so that I may respond `they'''.

What do we gain from composing the operations of linear pronoun logic?
Notably, they allow us to communicate that one pronoun is mandatory and another may be added on optionally.
If you \emph{must} call me \pron{she/her} and are \emph{welcome} to additionally sprinkle in \pron{they/them}, this can be captured by $\pron{she/her}\ \&\ (\pron{she/her} \otimes \pron{they/them})$.
Conversely, if I dictate the sprinkling of $\pron{they/them},$ this scenario would be expressed by $\pron{she/her} \oplus (\pron{she/her} \otimes \pron{they/them})$.

In scenarios such as this, one might perform a proof in linear logic to confirm whether a given protocol is safe!
One might prove $\pron{she/her} \multimap (\pron{she/her} \oplus (\pron{she/her} \otimes \pron{they/them}))$ as a means of confirming to oneself: ``no matter whether I have permission to use \pron{they/them} today or not, I will always be able to use \pron{she/her} in both cases!''

\section{Temporal Logic} 
Temporal logics are a family of modal logics whose modal operators express the truth of properties across a period of time, as opposed to eternal truth or truth only in some ``current'' moment.
Linear-temporal\footnote{The name of linear-temporal logic is fully unrelated to linear logic. The linear in linear-temporal logic refers to linear model of time, while in linear logic it refers to the usage patterns of assumptions.} logic (LTL)~\cite{pnueli1977temporal} is a particularly common version of temporal logic, so we take it as our basis here.

As with linear logic, the proposal of temporal logic as one potential foundation for pronoun logic is drawn directly from experiences in the authors' lives.
Even TGNB-affirming people frequently make incorrect assumptions about a stranger's gender, so it is not uncommon to correct a stranger's pronoun usage and then be met with a polite apology.
When this occurs, some speakers apologize so strongly as to violate the referent's wishes, e.g., they might continue apologizing long after referent wants the conversation to move on.
Engaging with this experience in an act of play, the authors propose: what if we could spell out which sequences of events over time are acceptable to us? 
Could we specify when it's okay to make an initial mistake and correct oneself, so that trans-affirming speakers might not descend into panic?
Linear-temporal (pronoun) logic is well-suited for this task, we wish to argue.

We present a proposed (fragment of)\footnote{The simplifications we make are substantial enough that they may throw off some experts in temporal logic; these are intentional simplifications due to reasons of audience.} linear-temporal pronoun logic, where the primary syntactic class (formulas) stands for pronoun descriptors (as a series of utilized pronoun classes by the speaker for the referent starting at first reference).
We define pronoun descriptors $A,B,C$ by the grammar:

\[ A,B,C {:}{:}{=}  s/o ~|~ \Box A ~|~ \Diamond A ~|~ A\ \mathcal{U}\ B ~|~ \bigcirc A~|~\cdots\]

We describe each construct in turn.
As in linear pronoun logic,  subject-object descriptors $s/o$ are the counterpart of propositional letters, standing in for a specific concrete pronoun set.
The box modality $\Box A$ means that the pronoun descriptor $A$ must be respected \emph{at all times}. This corresponds directly to the box modality of LTL, where $\Box A$ means a formula $A$ must hold true at all times now and in the future.
Its dual is the diamond modality $\Diamond A,$ meaning $A$ must be respected at \emph{some} (current or) future time, but not necessarily all. 
This again corresponds to the same modality in LTL, where $\Diamond A$ means that $A$ must be true either now or at some time in the future.
Because linear-temporal pronoun logic, like LTL, uses a discrete model of time, it is sensible to speak of ``the next moment,'' or more accurately, ``the next utterance.''
We write $\bigcirc A$ to mean that the next utterance must respect pronoun descriptor $A$.
This corresponds directly to the operator $\bigcirc A$ in LTL, meaning that formula $A$ must be true in the next instant.
The until modality $A\ \mathcal{U}\ B$ asserts that the pronoun descriptor $A$ might be in use for this utterance and some number of consecutive utterances into the future, there is some time in the future (or now) such that $B$ will be used. This is the same as in LTL, where $A\ \mathcal{U}\ B$ means that $A$ might hold true now and for some time into the future, but eventually $B$ will hold true at which point $A$ may or may not hold true.
We leave the definition open-ended with $\cdots$ because LTL typically includes propositional connectives such as negation $(\neg),$ conjunction $(\wedge),$ and disjunction $(\vee),$ but the exact choice of propositional primitives is uninteresting.
Notable definable operators include the bounded modalities $\Box_k A$ and $\Diamond_k A$ which respectively mean $A$ must be respected for all of the next $k$ utterances or that it must be respected somewhere within the next $k$ utterances.

Let us now explore the applications of these connectives to pronoun specification.
The descriptor $\Box she/her$ enforces a rigid protocol: one must always use she/her pronouns, and even a single mistake is a course for serious apology.
To the authors' best knowledge, this protocol is most widely employed by cisgender people, who have never had to experience misgendering on any frequent basis.
Conversely, the descriptor $\neg \pron{she/her}\ \mathcal{U}\ \Box\Diamond\pron{she/her}$\footnote{While $\neg \pron{she/her}\ \mathcal{U}\ \Box\Diamond\pron{she/her}$ is \emph{technically} logically equivalent to $\Box\Diamond\pron{she/her}$, this is for the purposes of self-expression and therefore the use of $\mathcal{U}$ is here in large part to denote a change in attitude.} is relatively permissive, accepting misgendering as long as the speaker begins using \pron{she/her} (at least sometimes) eventually (usually once corrected).

The compositional power of logic, however, lies in building up more complex specifications from simple ones.
In doing so, linear-temporal pronoun logic allows a high level of specificity.
Though high specificity might not be commonly seen in-the-wild in pronoun descriptors, this high level of specificity can be seen \emph{after-the-fact}, such as when a TGNB person privately unpacks a misgendering experience with a close friend, carefully deciding whether they felt respected or not, and why or why not.
Did they take two tries or five tries before respecting your pronouns?
That's the difference between $\Diamond_2 \pron{she/her}$ or $\Diamond_5 \pron{she/her}$.
Do you only tolerate the initial mistake if the fix is prompt?
That's $\neg A \wedge \bigcirc A$.
Do you only tolerate the mistake if the correct behavior continues in perpetuity?
That's $\neg A \wedge \bigcirc \Box A$
This example shows how compositions of temporal operators might arise in practice.

Other uses can be imagined as well. 
Perhaps a person uses dual $\pron{they/she}$ pronouns and has a preference for $\pron{they/them}$. 
They might half-jokingly announce ``if you never call me \pron{they} then you had better never call me \pron{he} either!''
This announcement could be captured by another compound specification: $\Box (A \vee \pron{they/them}) \vee (\Box \neg \pron{they/them} \wedge \neg \pron{he/him})$.
One might even draw inspiration from more established applications of LTL.
Liveness properties of the form $\Box \Diamond P$ are common: it should always be the case that correct behavior $P$ eventually occurs again.
Likewise $\Box \Diamond \pron{they/them}$ could indicate that one expects their correct pronoun to get used repeatedly throughout the future, no matter how often mistakes may occur.
Moreover, applications of LTL to safety-critical systems frequently employ reach-avoid specifications $\Box \neg A \land \Diamond B$.
The same pattern could be used to write a variant of previous half-joke, which perhaps more closely models the speaker's intentions: $\Box \neg \pron{he/him} \land \Diamond \pron{they/them}$.

\section{Description Operations in Free Logic} 
The definite and indefinite description operators $\iota x~p(x)$ and $\epsilon x~p(x)$ are logical formalizations of the natural-language determiners \emph{the} and \emph{some} respectively.
That is, $\iota x~p(x)$ is read as ``the unique $x$ satisfying $p(x)$'' and $\epsilon x~p(x)$ is read as ``some $x$ satisfying $p(x)$.''
Descriptions are often but not always explored in the context of \emph{free logic}~\cite{sep-logic-free}, a style of logic in which one may write terms which do not denote (mean) anything.
Beyond linguistic interest, free logic has found application for its flexibility, e.g., as an extensibility mechanism for interactive theorem-proving~\cite{DBLP:books/sp/NipkowPW02,DBLP:conf/cade/BohrerFP19}.
In general, the benefit of description operations is that they ensure a logic's \emph{term language} is as powerful as its \emph{formula language} because it allows importing formulas into terms.
We propose a free logic for \emph{genders} as opposed to pronouns, where a term describes a person, typically using one of the description operators.

We draw motivation again from the authors' lived experience.
Within TGNB communities, playing with ideas of gender is a common form of social play.
For some, this includes the production of \emph{microlabels}, identity labels which describe highly precise experiences.
In the extreme, microlabels include \emph{name-genders}, wherein a person employs their name in a gender label, capturing the individual uniqueness of their experience.
A less extreme example would be \emph{autigender}, wherein lived experiences of autism are deemed inseparable from the rest of an autistic person's gender identity.
Between TGNB people, one's gender identity could be an hours-long conversation rather than a simple ternary label ``woman, non-binary, or man''.
We propose description operations as a means to approximate the in-depth one-off description of one's gender experience in logic.
The complexity of descriptions one can create is dependent on the complexity of the predicates and functions we allow ourselves to use in a term.
In a practical implementation, one might need to settle on a specific set; here, we allow ourselves to use whichever predicates and functions we find convenient.


We outline  the proposal for free gender logic\footnote{Not to be confused with ``free genders!,'' itself an appealing proposal to the authors.} by giving a partial grammar for gender descriptors $A,B,C$
\[ A,B,C {:}{:}{=}   \iota x~A~|~ \epsilon x~A~|~x~|~p(x)~|~\forall x~ A ~|~ \cdots \] 
We leave the definition open-ended ($\cdots$) in a recognition that our free logic is a first-order logic and various definitions of first-order logic present differing equivalent sets of connectives.
The defining\footnote{Pun intended.} connectives are the definite and indefinite descriptions $\iota x~A$ and  $\epsilon x~A.$ 
Definite descriptions only denote when there exists exactly one $x$ satisfying $A$; their denotation is that $x$.
Indefinite descriptions only denote if there exists at least one $x$ satisfying $A$; they denote a fixed but arbitrary such $x$.
Variables $x,$ predicates $p(x),$ and quantifiers $\forall x~A$ are as is standard for free logics.

We now give example genders using our logic.
Some nonbinary people do not feel the need for a more precise label and are comfortable with self-describing just as ``neither male nor female.'' This is written $\epsilon x~(x \notin \mathsf{Males} \land x \notin \mathsf{Females}),$ i.e., this person is \emph{some} person who is neither male nor female.
During gender exploration, one might be more confident in what they \emph{are not} than in what they \emph{are}. The description $\epsilon x~ (x \notin \textsf{Females})$ means ``I might not know what I am, but it isn't a woman.``
Because free logic allows terms with no denotation (meaning) at all, it is also suitable for describing \emph{agender} people, people with no gender at all.
This can be done with $\epsilon$ using an always-false condition or with $\iota$ using a condition that is true for multiple values.
An example of the former is $\epsilon x~\textsf{false}$ and an example of the latter is $\iota x~(x \in \textsf{Males})$ because, although men exist, there are multiple of them.

We now return to the earlier motivating examples of microlabels such as autigender and name-genders.
One potential description of autigender follows:
\begin{align*}
\epsilon x~ &\mathsf{autistic}(x) \land \mathsf{nonbinary}(x) \\
            &\mathsf{important{-}to}(\mathsf{autism{-}of}(x),\mathsf{gender{-}of}(x))
\end{align*}
meaning that a person is autistic, nonbinary, and finds their autism important to their gender.
Name-genders are an example where the distinction between definite and indefinite description because essential.
Name-genders are typically meant to define a singular individual, thus definite description would be used, for example $\iota x~(\mathsf{name}(x) = \mathsf{Steve})$ for ``the one person named Steve.''
This illustrates a key challenge of name-genders: when many people have the same gender, names cannot be used to describe a unique person.
Definite description captures this accurately because if the person named Steve is not unique within the group of people under consideration (e.g., Steve's discourse community) then definite description accurately refuses to assign this term a meaning.
If one truly wishes for a gender that captures all Steves, they could use an indefinite description: $\epsilon x~(\mathsf{name}(x) = \mathsf{Steve})$.

How much modeling power does one truly gain with the introduction of description operators?
Ultimately, much of their benefit lies in their capacity to internalize the language of propositions within the language of terms, thus the gains are greatest in contexts where the term language is otherwise much weaker than the proposition language~\cite{DBLP:conf/cade/BohrerFP19}.
As we are exploring the question of logics for pronouns and gender from first principles without any established predecessor, such an evaluation would not be meaningful.
Instead, the benefit of descriptions is that they help us move beyond logics of pronouns into logics of gender.
Pronouns are a part of speech, and the previously proposed foundations of linear logic and linear-temporal logic are effective for specifying structured patterns of speech.

However, gender is more than pronouns, and a natural criticism of any formalism of gendered speech is that it will always be incomplete.
If we have formalized pronouns, one must not forget gendered occupational titles.
But we must also not forget gendered terms of endearment, compliments, or for that matter gendered vocal inflection, all of these are part of gendered language.
And gender goes far beyond language, potentially incorporating everything from one's style of dress to the internal sense of self.
It is unrealistic to expect such nuanced and individual experiences to be fully captured in any formalism, but the value of free gender logic is in demonstrating that formalism can do better than simply pronoun patterns in speech.
One can define a thing formally by defining its properties.
There is nothing stopping us from formally specifying properties of our gender experience that matter to us beyond pronouns.

\section{Related work}
Formalization of language, including pronouns, has long been an active topic of study in linguistics, both computational and otherwise.
Notable topics in formalization of pronouns include the relationship between pronouns and anaphora~\cite{jacobson2003binding}.
However, linguists more often seek to formalize the structure of natural language itself; our goal, in contrast, is to formalize a person's wishes regarding how language is used to address them.
Some researchers have pursued practical applications of formal modeling of pronouns, such as automated refactoring of sentences for gender-neutrality~\cite{sun2021they}.
This lends credence to our assertion that ``spell-checking'' is a reasonable direction for future applied work building on this paper.

The study of personal pronouns often intersects with the field of lavender linguistics, e.g., the study of language in LGBT+ communities.
Entire conferences are dedicated to this field and thus a comprehensive overview is not possible here.
We refer readers to Baron for accessible introduction to the history of personal pronouns, especially neopronouns, in relation to identity formation and civil rights advocacy~\cite{baron2020s}.
This work highlights the societal importance of gender pronouns in relation to gender identity, a core point in lavender linguistics:
```What's your pronoun?' is an invitation to declare, to honor, or to reject, not just a pronoun but a gender identity. And it's a question about a part of speech. Repeat: a question about a part of speech.''
In other words, because gender is a social construct, linguistic social processes serve a central whether a TGNB person feels accepted or rejected for who they are.
By building up new means of expressing our wishes for how these social processes play out, we can both empower those who wish to make us feel accepted and shed light on those who do not.

It is a common misconception that inclusive pronoun use, particularly the use of gender-neutral pronouns, is an inherently modern and inherently Western phenomenon.
It is certainly true that the visibility of TGNB people is increasing in the USA context~\cite{pewReport} and that English-language literature has given particular attention to gender-neutral pronouns in English (e.g., efforts at institutional change at Oxford~\cite{borza2021shall} and other European languages such as Swedish~\cite{gustafsson2015introducing}. 
A multi-lingual perspective complicates this narrative, however. 
In Mandarin Chinese, for example, gendered pronouns only emerged in 20th-century Republican-era language reforms~\cite{jortay2024gendered} in a direct emulation of other (European) languages, following a long history of gender-neutral pronouns.
What is true, however, is that the approach presented in our paper is inherently an approach for English, because the grammars of different languages create different real-world concerns for speakers and thus different needs and wants.
For example, a Japanese speaker, having gendered first-person pronouns, might care how another speaker reacts to the use of a certain first-person pronoun, something which does not apply in English.

\section{Practical Applications}
We caution that practical applications of pronoun formalizations need to be grounded in solid human-subjects research on the needs of TGNB users as a whole.
We assert that such studies are beyond our present scope, which focuses on critique, provocation, and play.

Nonetheless, we do not wish to ignore the likely practical applications.
By formalizing the meaning of PDs, one makes it possible to employ them in natural-language processing applications.
Spell-checkers could be developed which allow writers to detect misgendering in a draft and correct it before publication.
The same tools could be used, e.g., on a corpus drawn from social media, to study the prevalence of (intentional) misgendering in the wild, studies which might be used to advocate against hate speech more broadly.
While not applicable in languages such as English, which lack gendered second person pronouns, the usage of second person pronouns has been shown to improve the ability for students to absorb material in classrooms~\cite{cunningham2023put}. When translating for these languages and building computer systems for presenting coursework to students learning in said languages, the use of PDs would likely aid in the learning process.

\section{Conclusion}
This paper explored the ideas of Pronoun Logic and logics of gender more broadly through three proposals rooted respectively in linear logic, linear-temporal logic, and free logic with definite and indefinite descriptions.
Pronoun Logic refers to the use of formal logic to characterize a person's wishes regarding how others gender them when using pronouns in speech, while gender logic refers to logic for formally specifying broader aspects of gender identity, separately from pronoun usage.

Hypothetical applications of such work were laid out, such as checking a document's compliance with stated wishes and repairing it for compliance.
Our immediate goal, however, is to provoke reflection on the relationship between mathematical formalism and gender.
Traditional accounts in queer theory have deconstructed gender, resisting simplistic cisheteronormative conceptions of a gender binary.
For this reason, it is natural to remain skeptical of formalism. 
After all, gender is nuanced, individual, relational, and experiential: how could a formal definition capture it?

This perspective misses something essential: that our work in logic is also a part of who we are; if one closes off all use of mathematical modeling from experience of gender, then one closes off a part of us.
Mathematical formalisms, like gender, are socially constructed, and their construction is an act of creation.
We close in calling not for a rigid fixed formalization of gender pronouns in logic but for embracing mathematical play for play's sake, embracing the creative generation of new formalisms.

In so doing, we can do more than make a small group of TGNB researchers happy.  
By affirming that one's TGNB identity need to be separated from an interest in formalism, we also create important opportunities for outreach.
Young TGNB students with an interest in mathematics may be more open to pursuing higher education in these topics when they are presented in a light that celebrates those students' identities.
For this reason we propose future work developing high-school-level outreach materials based on pronoun logic, to show TGNB students that logic can be for them\footnote{We thank the anonymous reviewer who suggested this future work.}.
As they grow and their self-understanding deepens, they could be the ones who rewrite our logical foundations to encompass their diverse experiences of our world.

This fundamentally fluid, human-centered view of formalism is anything but the gender binary against which contemporary queer theory fought.

 \bibliography{paper}

\begin{thebibliography}{15}
\expandafter\ifx\csname natexlab\endcsname\relax\def\natexlab#1{#1}\fi

\bibitem[{Baron(2020)}]{baron2020s}
Dennis Baron. 2020.
\newblock \emph{What's your pronoun?: Beyond he and she}.
\newblock Liveright Publishing.

\bibitem[{Bohrer et~al.(2019)Bohrer, Fern{\'{a}}ndez, and
  Platzer}]{DBLP:conf/cade/BohrerFP19}
Rose Bohrer, Manuel Fern{\'{a}}ndez, and Andr{\'{e}} Platzer. 2019.
\newblock \href {https://doi.org/10.1007/978-3-030-29436-6\_6}
  {dl\({}_{\mbox{{\(\iota\)}}}\): Definite descriptions in differential dynamic
  logic}.
\newblock In \emph{Automated Deduction - {CADE} 27 - 27th International
  Conference on Automated Deduction, Natal, Brazil, August 27-30, 2019,
  Proceedings}, volume 11716 of \emph{Lecture Notes in Computer Science}, pages
  94--110. Springer.

\bibitem[{Borza(2021)}]{borza2021shall}
Natalia Borza. 2021.
\newblock Why shall i call you ze? discourse analysis of the social perception
  of institutionally introducing the gender-neutral pronoun ze.
\newblock \emph{Linguistik online}, 106(1):19--45.

\bibitem[{Cunningham et~al.(2023)Cunningham, Ahmed, March, Golden, Wilks, Ross,
  and McLean}]{cunningham2023put}
Sheila Cunningham, Zahra Ahmed, Joshua March, Karen Golden, Charlotte Wilks,
  Josephine Ross, and Janet McLean. 2023.
\newblock \href
  {https://journals.sagepub.com/doi/full/10.1177/17470218231174229} {Put you in
  the problem: Effects of self-pronouns on mathematical problem-solving}.
\newblock pages 308--325. Sage.

\bibitem[{DeYoung et~al.(2012)DeYoung, Caires, Pfenning, and
  Toninho}]{DBLP:conf/csl/DeYoungCPT12}
Henry DeYoung, Lu{\'{\i}}s Caires, Frank Pfenning, and Bernardo Toninho. 2012.
\newblock \href {https://doi.org/10.4230/LIPICS.CSL.2012.228} {Cut reduction in
  linear logic as asynchronous session-typed communication}.
\newblock In \emph{Computer Science Logic (CSL'12) - 26th International
  Workshop/21st Annual Conference of the EACSL, {CSL} 2012, September 3-6,
  2012, Fontainebleau, France}, volume~16 of \emph{LIPIcs}, pages 228--242.
  Schloss Dagstuhl - Leibniz-Zentrum f{\"{u}}r Informatik.

\bibitem[{Geiger and Graf(2019)}]{pewReport}
A.W.\ Geiger and Nikki Graf. 2019.
\newblock \href
  {https://www.pewresearch.org/short-reads/2019/09/05/gender-neutral-pronouns/}
  {About one-in-five u.s. adults know someone who goes by a gender-neutral
  pronoun}.

\bibitem[{Girard(1987)}]{DBLP:journals/tcs/Girard87}
Jean{-}Yves Girard. 1987.
\newblock \href {https://doi.org/10.1016/0304-3975(87)90045-4} {Linear logic}.
\newblock \emph{Theor. Comput. Sci.}, 50:1--102.

\bibitem[{Gustafsson~Send{\'e}n et~al.(2015)Gustafsson~Send{\'e}n, B{\"a}ck,
  and Lindqvist}]{gustafsson2015introducing}
Marie Gustafsson~Send{\'e}n, Emma~A B{\"a}ck, and Anna Lindqvist. 2015.
\newblock Introducing a gender-neutral pronoun in a natural gender language:
  the influence of time on attitudes and behavior.
\newblock \emph{Frontiers in psychology}, 6:145174.

\bibitem[{Jacobson(2003)}]{jacobson2003binding}
Pauline Jacobson. 2003.
\newblock Binding without pronouns (and pronouns without binding).
\newblock In \emph{Resource-sensitivity, binding and anaphora}, pages 57--96.
  Springer.

\bibitem[{Japaridze(1997)}]{DBLP:journals/apal/Japaridze97}
Giorgi Japaridze. 1997.
\newblock \href {https://doi.org/10.1016/S0168-0072(97)00046-8} {A constructive
  game semantics for the language of linear logic}.
\newblock \emph{Ann. Pure Appl. Log.}, 85(2):87--156.

\bibitem[{Jortay(2024)}]{jortay2024gendered}
Coraline Jortay. 2024.
\newblock Gendered language in modern chinese history.
\newblock In \emph{Routledge Handbook of Chinese Gender \& Sexuality}, pages
  42--55. Routledge.

\bibitem[{Nipkow et~al.(2002)Nipkow, Paulson, and
  Wenzel}]{DBLP:books/sp/NipkowPW02}
Tobias Nipkow, Lawrence~C. Paulson, and Markus Wenzel. 2002.
\newblock \href {https://doi.org/10.1007/3-540-45949-9} {\emph{Isabelle/HOL -
  {A} Proof Assistant for Higher-Order Logic}}, volume 2283 of \emph{Lecture
  Notes in Computer Science}.
\newblock Springer.

\bibitem[{Nolt(2021)}]{sep-logic-free}
John Nolt. 2021.
\newblock {Free Logic}.
\newblock In Edward~N. Zalta, editor, \emph{The {Stanford} Encyclopedia of
  Philosophy}, {F}all 2021 edition. Metaphysics Research Lab, Stanford
  University.

\bibitem[{Pnueli(1977)}]{pnueli1977temporal}
Amir Pnueli. 1977.
\newblock The temporal logic of programs.
\newblock In \emph{18th Annual Symposium on Foundations of Computer Science
  (sfcs 1977)}, pages 46--57. ieee.

\bibitem[{Sun et~al.(2021)Sun, Webster, Shah, Wang, and Johnson}]{sun2021they}
Tony Sun, Kellie Webster, Apu Shah, William~Yang Wang, and Melvin Johnson.
  2021.
\newblock They, them, theirs: Rewriting with gender-neutral {English}.
\newblock \emph{arXiv preprint arXiv:2102.06788}.

\end{thebibliography}

\end{document}